\documentclass[conference]{IEEEtran}
\IEEEoverridecommandlockouts

\usepackage{amsmath,amssymb,amsfonts}
\usepackage{algorithmic}
\usepackage{graphicx}
\usepackage{subcaption}
\usepackage{textcomp}
\usepackage{placeins}
\usepackage{hyperref}
\usepackage{booktabs}
\usepackage{cite}

\makeatletter
\newcommand{\linebreakand}{%
  \end{@IEEEauthorhalign}
  \hfill\mbox{}\par
  \mbox{}\hfill\begin{@IEEEauthorhalign}
}
\makeatother

\begin{document}
\title{UltraLight Med-Vision Mamba for Classification of Neoplastic Progression in Tubular Adenomas  \\

\thanks{The authors graciously thank the South Bend Medical Foundation (SBMF) for providing the whole slide images for this work.}
}

\author{\IEEEauthorblockN{Aqsa Sultana\IEEEauthorrefmark{1},
Nordin Abouzahra\IEEEauthorrefmark{1},
Ahmed Rahu, MD\IEEEauthorrefmark{2},
Brian Shula\IEEEauthorrefmark{3},\\
Brandon Combs\IEEEauthorrefmark{4},
Derrick Forchetti, MD\IEEEauthorrefmark{4},
Theus Aspiras, PhD\IEEEauthorrefmark{1},
Vijayan K. Asari, PhD\IEEEauthorrefmark{1}}
\IEEEauthorblockA{\IEEEauthorrefmark{1}\textit{Dept. of Electrical and Computer Engineering, University of Dayton, Dayton, OH USA}}
\IEEEauthorblockA{\IEEEauthorrefmark{2}
\textit{Dept. of Pathology, University of Toledo Medical Center, Toledo, OH, USA}}
\IEEEauthorblockA{\IEEEauthorrefmark{3}
\textit{Lead Mechanical Engineer, Honeywell International Inc., South Bend, IN, USA}}
\IEEEauthorblockA{\IEEEauthorrefmark{4}
\textit{Dept. of Pathology, South Bend Medical Foundation, South Bend, IN, USA}}
}

\maketitle

\begin{abstract}
Identification of precancerous polyps during routine colonoscopy screenings is vital for their excision, lowering the risk of developing colorectal cancer.  Advanced deep learning algorithms enable precise adenoma classification and stratification, improving risk assessment accuracy and enabling personalized surveillance protocols that optimize patient outcomes. UltraLight Med-Vision Mamba, a state-space-based model (SSM), has excelled in modeling long- and short-range dependencies and image generalization, critical factors for analyzing whole slide images. Furthermore, UltraLight Med-Vision Mamba’s efficient architecture offers advantages in both computational speed and scalability, making it a promising tool for real-time clinical deployment.
\end{abstract}


\section{Introduction}
Colorectal cancer (CRC) is a major global health challenge; in the United States, it’s the third most common cause of cancer and is the second leading cause of cancer-related death \cite{Siegel_Wagle_Cercek_Smith_Jemal_2023}. CRC frequently originates from colonic polyps, which are raised protrusions of colonic mucosa of epithelial origin, broadly categorized as adenomatous and serrated. Adenomatous polyps are due to neoplastic proliferation of glands and are a well-established, frequent precursor lesion to CRC \cite{Rosai_2011}. 
 For decades, CRC has been thought to arise through a traditional adenoma-carcinoma pathway \cite{df9de096f4994c3da5fb35aafa259138}, and screening guidelines have been established to detect precancerous lesions. The US Preventive Services Task Force recommendations show a substantial benefit to screening asymptomatic individuals starting at age 50 and a moderate benefit starting at age 45 \cite{Force_2021}.
Tubular adenomas, a subtype within the adenomatous polyps category, are one of such precancerous lesions, and are the primary focus of this discussion.

Tubular adenomas can be classified into two categories: those having low-grade dysplasia and those having high-grade dysplasia. The presence of high-grade dysplasia is a known risk factor for the development of CRC \cite{2581119744b04e758b2eecd8d5ecb83e} and represents an advanced stage along the adenoma-carcinoma continuum.
Despite prophylactic screening efforts, accurately assessing the malignant potential of low-grade tubular adenomas remains a clinical challenge. Traditional histopathological examination is based solely on visual assessment; the naked eye can face limitations in identifying subtle morphological features associated with patients’ long-term risks of developing subsequent CRC. 

The advent of digital pathology has enabled the generation of high-resolution whole slide images (WSIs). Combining WSIs with advancements in artificial intelligence (AI) has led to powerful new tools augmenting diagnostic accuracy and efficiency in pathology. This synergy between digital pathology and AI promises to improve the sensitivity of risk stratification and other aspects of clinical care that are impossible with traditional light microscopic examination alone\cite{korbar2017deeplearningclassificationcolorectalpolyps}. 

Bridging digital pathology alongside breakthroughs in deep learning offers unprecedented opportunities for objective and quantitative histological analysis. Deep learning models can now analyze, extract, and learn complex patterns directly from image data, identifying subtle morphological patterns and features imperceptible through visual assessment \cite{korbar2017deeplearningclassificationcolorectalpolyps}. This study implements Vision Mamba, \cite{wu2024ultralightvmunetparallelvision}, a novel State Space Model (SSM)   \cite{zhu2024visionmambaefficientvisual} architecture, to analyze intricate histological patterns within WSIs of low-grade tubular adenomas to identify subtle indicators associated with subsequent colorectal cancer risk. The model also leverages an ultra-light architecture preventing parameter explosion, making it a promising tool
for real-time clinical deployment \cite{zhu2024visionmambaefficientvisual}.

\section{Methodology}

\begin{figure*}[htbp]
    \centerline{%
        \includegraphics[width=1.05\linewidth,height=10cm]{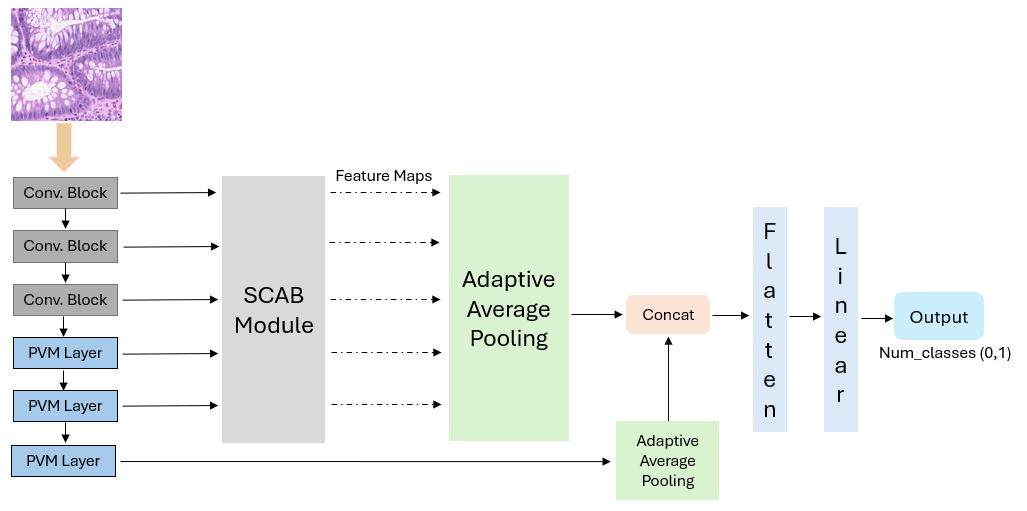}%
    }
    \caption{Architectural structure of UltraLight Med-Vision Mamba model for image classification task.}
    \label{fig1}
\end{figure*}

\begin{figure*}[htbp]
    \centerline{%
        \includegraphics[width=1.05\linewidth,height=8cm]{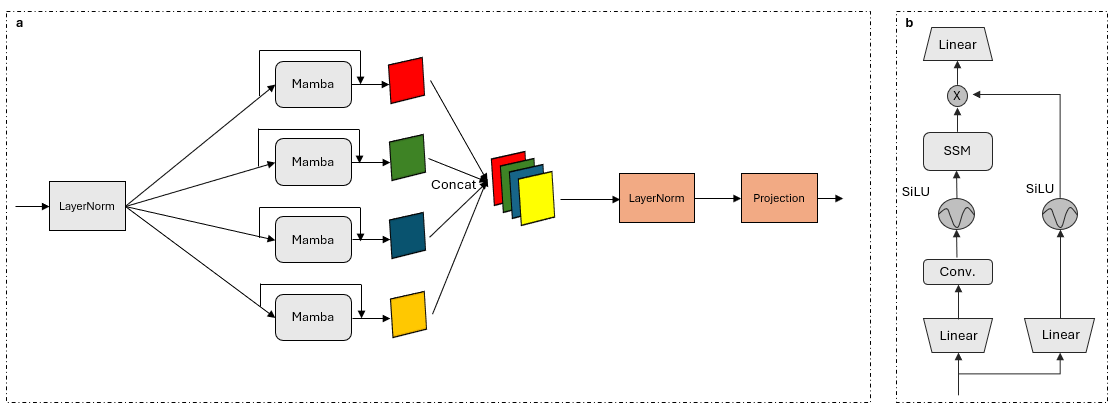}%
    }
    \caption{a) PVM layer in UltraLight Med-Vision Mamba b) Mamba module.}
    \label{fig2}
\end{figure*}

The UltraLight Med-Vision Mamba model \cite{Sultana_Abouzahra_Asari_Aspiras_Liu_Sudakow_Cooper_2025, wu2024ultralightvmunetparallelvision}  adopts an architectural framework akin to convolutional neural networks (CNNs), but with a key distinction: instead of relying on convolutional blocks as its primary feature extractors, it employs Parallel Vision Mamba (PVM) layers. The overall architecture comprises six layers, with the number of channels configured as \([8,16,24,32,48,64]\) as shown in Fig. \ref{fig1}. The initial three layers utilize standard convolutional blocks to extract shallow features, while the deeper layers (layers 4 through 6) incorporate PVM layers to capture more complex and nuanced features. The extracted features from three convolutional blocks and two PVM layers are fed into SCAB (spatial and channel attention bridge) module. Adaptive average pooling is performed on feature maps acquired from SCAB and the final PVM layer (stage 6) to standardize spatial dimensions of the feature maps. The pooled features are then concatenated to accumulate and retain the relevant information. The classification head first flattens the high-dimensional feature maps into a one-dimensional vector, effectively aggregating the spatially distributed information extracted by the preceding layers. This flattened vector is then passed through fully connected (dense) layers, which serve to map the learned feature representations to the final output space. In this case of colorectal adenoma classification, this typically corresponds to a set of class probabilities, enabling the model to assign a likelihood score to each potential category of control and cancer group.

The Parallel Vision Mamba (PVM) layer \cite{Sultana_Abouzahra_Asari_Aspiras_Liu_Sudakow_Cooper_2025, wu2024ultralightvmunetparallelvision}, as shown in Fig. \ref{fig2} (a), is also known as the PVM module. It incorporates Mamba blocks with residual connections from the input to the output of Mamba blocks to enhance the model’s ability to capture complex spatial relationships.
The input first undergoes layer normalization, after which the feature maps are split into four distinct branches, each with a designated number of channels. These branches are processed independently through the Mamba mechanism. The outputs from the Mamba blocks are then combined with residual connections from the original inputs, along with an adjustment factor to optimize learning. The resulting feature maps are concatenated to form four unified feature maps with specific channel dimensions. These concatenated outputs are subsequently normalized again and passed through a projection layer. By processing features in parallel across multiple branches, the PVM module is able to extract multiscale and intricate feature representations using varying kernel sizes. Moreover, this design efficiently reduces the number of parameters by preserving the same receptive field, thereby mitigating the parameter growth typically associated with increasing channel dimensions--an important consideration, as the parameter count in Mamba layers is highly sensitive to input channel size.

The model performance is further improved by the addition of SCAB module, also known as the Spatial and Channel Attention Bridge \cite{ruan2022malunetmultiattentionlightweightunet, wu2024ultralightvmunetparallelvision}, for feature propagation. Spatial attention bridge consists of max-pooling, average pooling, and extended convolution of shared weights. Channel attention bridge includes fully connected layers (FCL), global average pooling (GAP), concatenation, and sigmoid activation function. The SCAB module enhances the sensitivity, ability of the model to converge, and fusion of multi-scale features of different scales \cite{wu2024ultralightvmunetparallelvision,Sultana_Abouzahra_Asari_Aspiras_Liu_Sudakow_Cooper_2025}.

\section{Training and Experimental Results}
\subsection{Training Method}
The baseline for the experiment was established by training all models---Vision Transformer (ViT), swin Transformer, and UltraLight Med-Vision Mamba---for 100 epochs using Stochastic Gradient Descent (SGD) with a momentum of 0.9 and an initial learning rate of 0.001. 
UltraLight Med-Vision Mamba was further fine-tuned with a learning rate range of 0.0001 to 0.05 and trained for 300 epochs. Binary Crossentropy was used as the loss function. The training strategy incorporated the OneCycle Learning Rate (OneCycleLR) scheduler and Stochastic Weight Averaging (SWA) to stabilize training. The experiments were implemented using the PyTorch framework in Python on a NVIDIA GeForce RTX Titan GPU. 

\subsection{Transformer Based Models}
\noindent \textit{\textbf{Vision Transformer}} \cite{DBLP:journals/corr/abs-2010-11929}, also known as ViT, divides the input image into fixed-size patches, flattens them, and treats them like tokens in a sequence similar to Natural language Processing (NLP) \cite{vaswani2023attentionneed}. The patches are then processed using multi-self-attention-heads to capture global context. While powerful, ViT requires large datasets and high computational resources.\\

\noindent \textit{\textbf{Swin Transformer}} \cite{DBLP:journals/corr/abs-2103-14030} also known as \textbf{S}hifted \textbf{Win}dow Transformer is a more efficient variant of Transformers that hierarchically processes images using non-overlapping local windows. It introduces shifting window mechanism that allows cross-window connections while maintaining computational efficiency. This makes the Swin Transformer more scalable for smaller datasets and dense prediction tasks than ViT.

\subsection{Model Parameters}
The comparison between the number of model parameters for the three architectures is presented in Table \ref{tab:tab2}. ViT has the highest number of parameters at \(7,398,785\), indicating large model size and computational overhead. Swin Transformer, with its heirarchical architectural structure, is designed for more efficient computation significantly reduces the model parameter count to \(598,099\). Lastly, UltraLight Med-Vision Mamba, dramatically reduces the model parameter to only \(49,641\)---making it the most lightweight model among the three. This suggests that UltraLight Med-Vision Mamba is highly optimized for efficiency, better trade-off between performance and computational cost, especially for the real-time clinical deployment.
\begin{table}[h!]
    \begin{center}
        \caption{Model parameter comparison for ViT, Swin Transformer and UltraLight Med-Vision Mamba.}
        \label{tab:tab2}
        \begin{tabular}{lc}
            \toprule
            \textbf{Model} & \textbf{Model Parameters} \\
            \midrule
            ViT & 7,398,785 \\
            Swin Transformer & 598,099 \\
            \textbf{UltraLight Med-Vision Mamba} & \textbf{49,641} \\
            \bottomrule
        \end{tabular}
    \end{center}
\end{table}

\subsection{Dataset}
\begin{figure}[h]
    \centering 
    
     \begin{subfigure}[b]{0.22\textwidth}
        \captionsetup{labelformat=empty}
        \includegraphics[width=\linewidth]{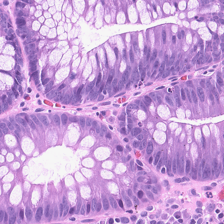}
        \caption{Case}
        
    \end{subfigure}
    \begin{subfigure}[b]{0.22\textwidth}
        \captionsetup{labelformat=empty}
        \includegraphics[width=\linewidth]{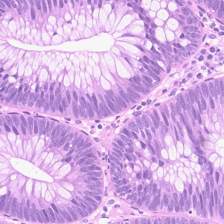}
        \caption{Control}
    \end{subfigure}

    \caption{Sample images of the dataset used: case group (left) and control group (right).}
    \label{fig:fig3}
\end{figure}

The original whole slide images (WSIs) were tiled at 1024 × 1024
pixel resolution, with 3 color channels and then resized into smaller tiles of 224 × 224 pixels with 3 color channels for model input as shown in Fig. \ref{fig:fig3}. During preprocessing, a region-of-interest (ROI) filter was applied to determine whether each tile should be retained or discarded. All automatically generated ROIs were subsequently subjected to visual inspection to verify annotation accuracy. Tiles exhibiting quality issues—such as tissue folding, edge artifacts, or poor scan resolution—were excluded through manual review of WSI patch location maps. After this curation process, a total of 176,945 high-quality tiles were retained in each class. The final dataset was divided into training (70\%), validation (15\%), and testing (15\%) subsets.

\subsubsection{Data Management Strategies}
Patients without known high-risk clinical factors for CRC, in which low-grade tubular adenomas were identified during screening colonoscopy, were included in the study. A total of 81 patients (41 male, 40 female), ranging in age from 54-95 years (average 70), underwent at least one screening colonoscopy with associated biopsies demonstrating tubular adenomas with low-grade dysplasia; no biopsies showed any histologic features that were indicative of high-risk progression to CRC. Patients were stratified into two cohorts: a precancer group and a control group. The case group consisted of individuals who subsequently developed CRC following screening colonoscopies in which low-grade tubular adenomas were identified. The control group comprised of individuals with no history of CRC despite having low-grade adenomas detected on one or more screening procedures. Compared to the case group, the control group had a greater average number of biopsies and a longer mean screening interval. On average, patients in the case group were 6.86 years older than those in the control group. Histologic slides from both groups containing low-grade tubular adenomas were digitized using the same Leica Aperio AT2 whole slide scanner to generate image data for this study.  

\subsection{Results}
Quantitative comparisons---Accuracy, F1, Precision and Recall---for colorectal adenoma classification using ViT, Swin Transformer, and UltraLight Med-Vision Mamba are summarized in Table \ref{tab:tab1}. While the transformer-based models (ViT and Swin Transformer) primarily utilize self-attention mechanisms to model long-range dependencies across image patches achieved comparable performance with accuracies of 89.84\% and 89.52\% respectively. In contrast, UltraLight Med-Vision Mamba employs State Space Models (SSMs), which process sequences bidirectionally. This approach allows UltraLight Med-Vision Mamba to achieve 97.34\% with higher F1, Precision and Recall, indicating balanced and robust performance. Its ability
to better capture subtle dependencies within high-resolution images offered it a distinct advantage. The incorporation of the SCAB module enhanced the feature propagation required for the image classification task. As the SSM models the hidden states over time, it offers the ability to excel in modeling long- and short-range dependencies.
\begin{table}[h!]
    \begin{center}
        \caption{Quantitative performances of ViT, Swin Transformer and UltraLight Med-Vision Mamba.}
        \label{tab:tab1}
        \resizebox{\linewidth}{!}{ 
        \begin{tabular}{lcccc}
            \toprule
            \textbf{Model} & \textbf{Accuracy} & \textbf{F1} & \textbf{Precision} & \textbf{Recall} \\
            \midrule
            Vision Transformer & 89.84\% & 0.8920 & 0.9519 & 0.8392 \\
            Swin Transformer & 89.52\% & 0.8878 & 0.9548 & 0.8296 \\
            \textbf{UltraLight Med-Vision Mamba} & \textbf{97.34\%} & \textbf{0.9733} & \textbf{0.9780} & \textbf{0.9686} \\
            \bottomrule
        \end{tabular}
        }
    \end{center}
\end{table}


\section{Discussion}

\begin{figure}
    \centering
    
    
    \centering
    \begin{subfigure}[b]{0.22\textwidth}
        \captionsetup{labelformat=empty}
        \includegraphics[width=\linewidth]{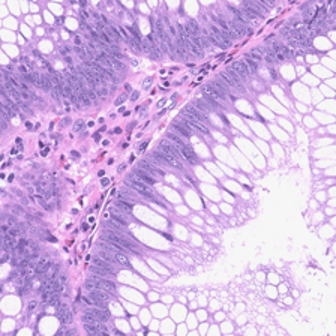}
        \caption{a.) Target: Case, Predicted: Control}
        
    \end{subfigure}
    \begin{subfigure}[b]{0.22\textwidth}
        \captionsetup{labelformat=empty}
        \includegraphics[width=\linewidth]{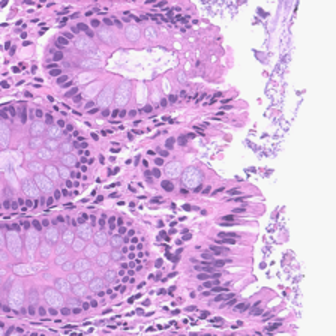}
        \caption{b.) Target: Control, Predicted: Case}
    \end{subfigure}
    \caption{Mismatched predicted outputs of UltraLight Med-Vision Mamba.}
    \label{fig:fig4}
  
    \hfill

    \vspace{0.1cm} 


    \centering
      \begin{subfigure}[b]{0.22\textwidth}
        \captionsetup{labelformat=empty}
        \includegraphics[width=\linewidth]{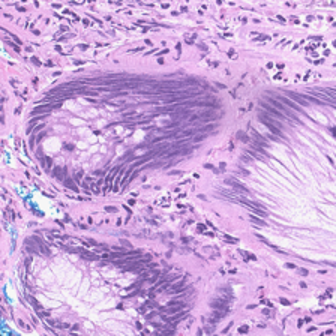}
        \caption{a.) Target: Case, Predicted: Control}
    \end{subfigure}
    \begin{subfigure}[b]{0.22\textwidth}
        \captionsetup{labelformat=empty}
        \includegraphics[width=\linewidth]{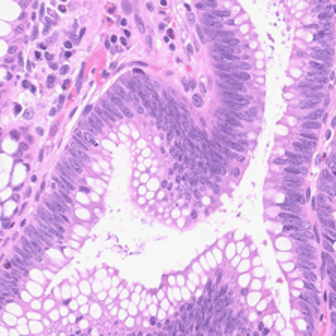}
        \caption{b.) Target: Control, Predicted: Case}
    \end{subfigure}
    \caption{Mismatched predicted outputs of Vision Transformer.}
    \label{fig:fig5}

    \vspace{0.1cm} 
   
    \centering
    \begin{subfigure}[b]{0.22\textwidth}
        \captionsetup{labelformat=empty}
        \includegraphics[width=\linewidth]{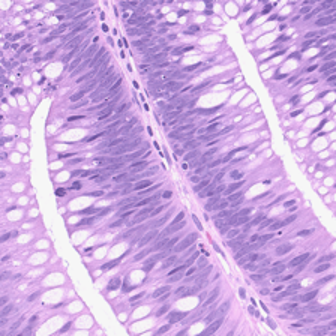}
        \caption{a.) Target: Case, Predicted: Control}
    \end{subfigure}
    \begin{subfigure}{0.22\textwidth}
        \captionsetup{labelformat=empty}
        \includegraphics[width=\linewidth]{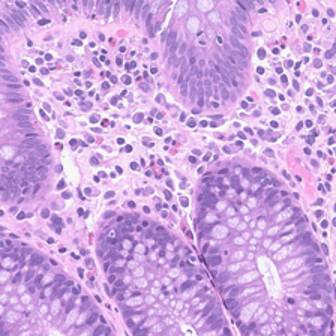}
        \caption{b.) Target: Control, Predicted: Case}
    \end{subfigure}
    \caption{Mismatched predicted outputs of Swin Transformer.}
    \label{fig:fig6}
\end{figure}

This section discusses the limitations of the model's ability to generalize and interpret in colorectal histopathology. The misclassified predictions from each of the three architectures--UltraLight Med-Vision Mamba, ViT, and Swin Transformer--are shown in Figures \ref{fig:fig4} to \ref{fig:fig6}. This illustrates the visual and histological features that may have contributed to incorrect classifications.  \\

\noindent \textbf{\textit{Predicted outputs of UltraLight Med-Vision Mamba:}} In Fig. \ref{fig:fig4} (a), the patch clearly demonstrates features of tubular adenoma: nuclear pseudostratification, hyperchromatic elongated nuclei, and goblet cell depletion. Despite the cytological hallmarks being present, the model misclassified this region as class control. This may perhaps reflect the model’s limited sensitivity to focal dysplasia, particularly in areas with mixed histology, where adjacent non-dysplastic crypts can mask subtle neoplastic changes.
These results may highlight the need for enhanced model training with finer-grained annotations and increased representation of early or borderline dysplasia cases. 

In Fig. \ref{fig:fig4} (b), the region is histologically consistent with slight adenomatous changes from the control group and was misclassified by the model as a tubular adenoma that progressed to CRC (case group). The crypts are well-formed, with preserved spacing, abundant goblet cells, and basally aligned nuclei, lacking pseudostratification or any cytologic atypia that may be indicative of a high-grade dysplasia, or any other signs of progression to CRC.  
This false positive may be due to subtle visual cues such as epithelial overcrowding near the tissue edge or darker nuclear staining, which the model may overfit during training.\\

\noindent \textbf{\textit{Predicted outputs of Vision Transformer:}} 
The Fig. \ref{fig:fig5} (a) is an image from the case group that was misclassified with a control group prediction. The model's benign interpretation was likely influenced by confounding factors such as tangential plane of sectioning and the absence of pronounced pseudostratification. The preserved goblet cells and rounded glandular contours may have biased the classifier towards a benign interpretation. This suggests that architectural orientation and crypt profile may significantly influence model sensitivity to dysplasia. 

 Figure \ref{fig:fig5} (b) was expected to be classified as control, but was misclassified as case by the model. The likely reason lies in the cytologic pseudostratification observed in the glandular epithelium, which mimics low-grade dysplasia. The elongated nuclei aligned perpendicularly to the basement membrane, as well as the increase in nuclear crowding, may have falsely signaled dysplastic changes in the classifier.\\

\noindent \textbf{\textit{Predicted outputs of Swin Transformer:}}

A false negative occurred for Fig. \ref{fig:fig6} (a), which represents a tubular adenoma from the case group, perhaps due to its bland architecture and relatively preserved nuclear polarity. While mild nuclear elongation as well as pseudostratification are indeed present, the retention of goblet cells and lack of architectural complexity may have masked dysplastic cues from the classifier. This highlights the challenge of detecting low-grade adenomas that closely mimic normal mucosa at a higher magnification.  

A false positive classification was made for Fig. \ref{fig:fig6} (b), originally belonging in the control group. The classifier likely over-weighted features of mild atypia associated with inflammation due to an influx of lymphocytes and plasma cells in one of the colonic layers. In the real-life practice of pathology, it is common for these reactive epithelial changes to mimic dysplastic characteristics from the case group, leading to erroneous prediction(s). This case highlights a critical challenge for classification of histopathological specimens: discerning true dysplastic (and in some applications, neoplastic) changes from a wide range of confounding inflammatory and reactive processes. 


\section{Conclusion}
This study demonstrates the potential of the UltraLight Med-Vision Mamba architecture for improving the classification of low-grade colorectal adenomas from whole slide images. By effectively modeling long and short-range dependencies and complex spatial relationships through Parallel UltraLight Med-Vision Mamba layers, the network captures subtle histological patterns by enhancing feature extraction that conventional methods may overlook.

\section*{Acknowledgment}
The authors would like to thank the South Bend Medical Foundation for generously providing the dataset and medical insight required in this study.

\nocite{*}
\bibliographystyle{IEEEtran}
\bibliography{NAECON-conference-paper}

\end{document}